\documentclass[dvipsnames,format=sigconf,anonymous=false,review=false,authordraft=false]{acmart}

\usepackage[ruled,vlined]{algorithm2e}

\begin{document}

\title{LLM Guided Evolution - The Automation of Models Advancing Models}

\author{Clint Morris, Michael Jurado, and Jason Zutty}
\email{clintmaxmorris@gmail.com, michael.jurado@gtri.gatech.edu, jason.zutty@gtri.gatech.edu}
\affiliation{%
  \institution{Georgia Tech Research Institute}
  \streetaddress{925 Dalney St NW}
  \city{Atlanta}
  \state{Georgia}
  \country{USA}
  \postcode{30332}
}

\acmConference[GECCO '24]{Genetic Evolutionary Computing Conference}{July 14--18, 2018}{Melbourne, Australia}

\begin{abstract}
In the realm of machine learning, traditional model development and automated approaches like AutoML typically rely on layers of abstraction, such as tree-based or Cartesian genetic programming. Our study introduces "Guided Evolution" (GE), a novel framework that diverges from these methods by utilizing Large Language Models (LLMs) to directly modify code. GE leverages LLMs for a more intelligent, supervised evolutionary process, guiding mutations and crossovers. Our unique "Evolution of Thought" (EoT) technique further enhances GE by enabling LLMs to reflect on and learn from the outcomes of previous mutations. This results in a self-sustaining feedback loop that augments decision-making in model evolution. GE maintains genetic diversity, crucial for evolutionary algorithms, by leveraging LLMs' capability to generate diverse responses from expertly crafted prompts and modulate model temperature. This not only accelerates the evolution process but also injects expert like creativity and insight into the process. Our application of GE in evolving the ExquisiteNetV2 model demonstrates its efficacy: the LLM-driven GE autonomously produced variants with improved accuracy, increasing from 92.52\% to 93.34\%, without compromising model compactness. This underscores the potential of LLMs to accelerate the traditional model design pipeline, enabling models to autonomously evolve and enhance their own designs. 

\end{abstract}




\maketitle

\section{Introduction}
In the ever-evolving domain of machine learning, the convergence of human cognitive skills and automated algorithms is entering a pivotal junction. This paper introduces “Guided Evolution” (GE), a novel framework that combines the human-like expertise of Large Language Models (LLMs) with the robust capabilities of Neural Architecture Search (NAS) through genetic algorithms. This innovative fusion advances automated machine learning, elevating traditional NAS by integrating a more insightful, intelligently guided evolutionary process.

Central to this framework is our  “Evolution of Thought” (EoT) technique, which extends and refines concepts like Zero-Shot Chain-of-Thought, Automated Chain-of-Thought, and Tree-of-Thought \cite{kojima2023large, zhang2023automatic, yao2023tree}. These methodologies aim to improve the reasoning capabilities of LLMs. EoT takes a unique step forward by enabling LLMs to receive result-driven feedback, empowering them to make informed improvements based on the performance of their prior code augmentations, a significant advancement in intelligent automated machine learning.

EoT catalyzes LLMs to introspect and fine-tune suggestions based on past iterations, creating a self-enhancing feedback loop that fine-tunes architectural evolution. At the same, GE maintains essential genetic diversity for evolutionary algorithms while injecting human-like expertise and creativity into the evolutionary framework. Building from the insights of Ma et al. \cite{ma2023conceptual}, our Guided Evolutionary framework is further enhanced by a Character Role Play (CRP) technique, to markedly increase the feasibility, usefulness and creativity of ideas engendered by the LLM. 

The effectiveness of the Guided Evolution (GE) framework is showcased in the evolution of the ExquisiteNetV2 model. This evolution, initiated with a State-Of-The-Art (SOTA) seed model, not only demonstrates the capacity of LLMs to build upon and enhance SOTA models in collaboration with human expertise but also underscores their autonomous model design. This case study illustrates the framework's self-sufficient ability to generate improved model variants, emphasizing the burgeoning impact of LLMs in redefining traditional model design pipelines, a step towards models that independently evolve and refine their architectures.

\section{Neural Architecture Search}
Neural Architecture Search (NAS) stands at the forefront of machine learning innovation, focusing on the automated discovery of optimal neural network architectures. Within this dynamic field, a variety of methodologies have emerged, each contributing unique approaches and benefits. These include Reinforcement Learning (RL), Evolutionary Algorithms (EAs), Surrogate Model-Based Optimization, and One-Shot Architecture Search.

Focusing first on Reinforcement Learning, important contributions by Zoph and Le \cite{zoph2016neural} and Baker et al. \cite{baker2017accelerating} have showcased how an RL agent can be used to iteratively optimize network architectures, with an emphasis on maximizing key performance metrics like validation accuracy. Building on this foundation, Ramachandran, Zoph, and Le expanded the RL \cite{ramachandran2017searching} application to include the search for activation functions, furthering the versatility of this approach.

In a distinct but equally innovative vein, Evolutionary Algorithms draw inspiration from the principles of biological evolution. They utilize mutation and crossover processes to evolve network architectures. The potential of neuroevolution in NAS was aptly demonstrated by Real et al. \cite{real2019regularized}, creating the first evolved model to surpass hand-designs on ImageNet. Complementing this, CoDeepNEAT, developed by Miikkulainen et al. \cite{miikkulainen2017evolving}, extends the NEAT algorithm to deep learning, co-evolving modules and blueprints for constructing deep neural networks, thus highlighting the adaptability of evolutionary approaches in NAS. Furthermore, Lu et al. \cite{lu2019nsga} introduced NSGA-NET, a multi-objective genetic algorithm, signifying a substantial leap forward in this domain. 

Surrogate Model-Based Optimization diverges from RL's trial-and-error approach by employing predictive models to estimate the performance of various architectures to better navigate the architecture space Cai et al. \cite{cai2018path} .

Lastly, the realm of One-Shot Architecture Search represents a paradigm shift in NAS. It involves training an expansive super network that encompasses all potential architectures within the search space. Notable contributions in this area include the Efficient Neural Architecture Search (ENAS) by Pham et al. \cite{pham2018efficient}, and the Differentiable Architecture Search (DARTS) by Liu, Simonyan, and Yang \cite{liu2018darts}. By leveraging shared weights across architectures, these methods significantly reduce computational requirements.

In the rapidly evolving field of machine learning, our research introduces novel methodologies that synergize evolutionary algorithms (EA) with the reasoning and generative capabilities  of LLMs. We present "Guided Evolution" and "Evolution of Thought" (EoT), innovative approaches designed to address the inherent inefficiencies in traditional EA, a challenge highlighted in the works by Guariso et al. \cite{guariso2020improving} and Yang et al. \cite{yang2022improved}. These inefficiencies become particularly pronounced with the increasing size and training costs of modern machine learning models, necessitating a novel approach to enhance efficiency and adaptability.

Our methodologies integrate LLMs within the genetic algorithm framework, thereby introducing a robust intrinsic feedback mechanism (EoT) and an intelligently supervised evolution, where mutations and genetic crossovers are chosen by the LLMs framework. This integration not only alleviates the inefficiency issues but also significantly reduces the manual intervention often required in EA, such as parameter tuning and range setting. By leveraging the domain expertise of LLMs, we achieve a dynamic, efficient, and adaptable framework for exploring complex solution spaces in machine learning. Moreover, genetic diversity is preserved through the LLM's capability to produce an unlimited number of responses from a single prompt. This is achieved by adjusting the temperature parameter and utilizing a range of hand crafted prompts, thereby facilitating a thorough exploration of alternative model designs.

Traditional EA methods are often limited by their rigid mutation and crossover structures, a constraint partially addressed by developments such as those by Zutty et al. \cite{zutty2015multiple}, who introduced human-derived primitives. However, the utilization of predefined building blocks and tree structures in traditional genetic programming necessitates a prolonged set-up time. Rather, by generating individuals directly as interpretable code, LLMs save time by automatically enriching the evolutionary process with contextual understanding and nuanced insights which normally would come from human-derived primitives. Moreover, they are not constrained by a rigid structure; their development is limited only by the LLM's creativity.

The prevalent issue with LLMs stems from their design to emulate, rather than authentically generate, human-like responses. This often results in the production of erroneous or fabricated information, a phenomenon known as "hallucination" \cite{huang2023survey}. Addressing this, prompt engineering emerges as a method to navigate LLMs towards structured, logical reasoning via carefully formulated prompts. Our research significantly extends this nascent field. Additionally, an advantage of evolution is in the natural experimentation that arises during the process: a "hallucinated" piece of code will be evaluated against supervised data, its objectives and fitness scores accurately reflecting its success. Not every individual need be a winner as long as the evolution continues to make steady progress.

Building on the foundational work of Kojima et al. \cite{kojima2023large} in Chain-of-Thought (CoT) reasoning and the advancements by Zhang et al. \cite{zhang2023automatic} in Auto-CoT, our study introduces the EoT approach. This method incorporates genetic algorithms to enable LLMs to autonomously curate and refine prompts. Detailed in Section \ref{sec:EoT}, EoT leverages selective evolutionary strategies to not only augment the quality of LLM outputs but also to introduce a mechanism for self-optimization. This approach marks a step forward in developing self-improving, adaptive language models, showcasing the potential for more autonomous and efficient evolution in the field of machine learning.

\section{Methodology}
\label{sec:Methodology}

In our methodology, we introduce the Guided Evolution (GE) framework, initiating with ExquisiteNetV2 as the seed model. This model is first dissected into discrete code blocks, each corresponding to a distinct Python class. These blocks function analogously to genetic segments in a genome, providing the foundational elements for our LLM driven evolutionary process.

For this study, we utilized Mixtral \cite{jiang2024mixtral}, Mistral AI’s 8x7B Mixture of Experts Open Source Model, noted for its balance of efficiency and high performance in code generation. Mixtral employs a Sparse Mixture of Experts (SMoE) architecture, selectively utilizing 13B out of 47B parameters across eight feedforward blocks for each token, optimizing for inference efficiency. This model demonstrates superior performance in code generation compared to competitors like Llama 2 70B, making it particularly effective for the precise and complex code modifications required in our Guided Evolutionary framework. 

In a departure from traditional evolutionary algorithms (EA), our approach replaces conventional mutation and mating operations with a series of prompts directed at a LLM. This introduces a more dynamic and exploratory dimension to the EA framework. To further encourage innovative architectural evolution, we utilize "Character Role Play" (CRP), as outlined in Section \ref{sec:character}. This strategy enhances the creativity and unconventionality of guided mutations.

Another key feature of our methodology is the EoT approach (detailed in Section \ref{sec:EoT}). This induces an intrinsic feedback mechanism within the EA framework, allowing it to adapt and respond to effective changes in the model architecture. Through this mechanism, the LLM becomes attuned to successful adaptations, guiding the evolution of the model architecture in a more informed and targeted manner.

This combination of LLM repeated prompting and the EoT methodology creates a dynamic environment for the guided evolution of the seed model. It allows for both exploratory diversification and the exploitation of effective architectural changes, providing a novel approach to evolving neural network architectures.

By breaking down the ExquisiteNetV2 model into modular segments, such as the feature concentrator, we can direct the LLM's capabilities more precisely. This focused approach ensures that each segment is optimized in isolation, allowing for a detailed and specific enhancement of the model's overall architecture. Table \ref{tab:model_decomposition} delineates the decomposition of the state-of-the-art model into distinct code segments, providing a structured overview of its components.

\begin{table}
    \centering
    \begin{tabular}{p{3.2cm} p{4cm}}
        \hline
        \multicolumn{1}{c}{\textbf{Gene Segment}} & \multicolumn{1}{c}{\textbf{Description}} \\
        \hline
        get\_optimizer & Optimization Process \\
        SE & Squeeze-and-Excitation \\
        SE\_LN & Squeeze-and-Excitation Layer-Normalization\\
        DFSEBV2 & Feature-Processor \\
        FCT & Feature-Concentrator \\
        EVE & Extreme-Value-Expansion \\
        ME & Max-Min Expansion \\
        DW & Depthwise Convolution \\
        ExquisiteNetV2 & Aggregator \\
        \hline
    \end{tabular}
    \caption{Block Descriptions}
    \label{tab:model_decomposition}
\end{table}

\subsection{Mating}
\label{sec:mating}

\begin{figure}
    \centering
    \includegraphics[width=1\linewidth]{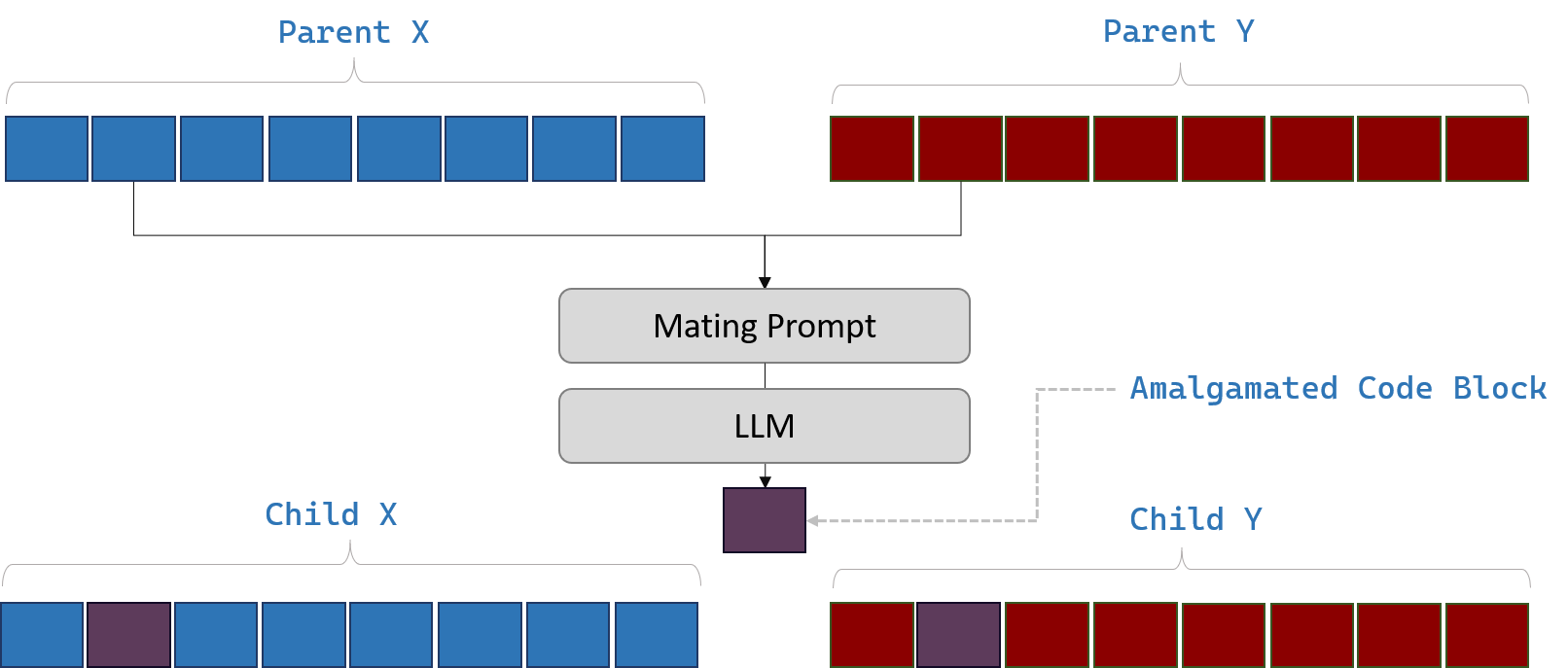}
    \caption{LLM Driven Code Block Mating}
    \label{fig:llm_mating}
\end{figure}

To enhance genome mating efficiency, the Guided Evolution (GE) framework employs a strategic selection process. Figure \ref{fig:llm_mating} shows the process of randomly selecting two distinct code segments from the available genomes, ensuring they are not identical to avoid ineffectual mating attempts. These selected segments are then processed through a LLM. The LLM's task is to intelligently amalgamate these segments, aiming to either heighten accuracy or boost efficiency in the resultant genome.

\subsection{Mutation}
\label{sec:mutation}

Incorporating a LLM into our mutation process has introduced a higher degree of flexibility within the Guided Evolutionary framework. Mutation here involves the random selection of code segments, which are then augmented through LLM prompts. To enhance the diversity of these mutations, we varied the model temperature for each prompt between 0.05 and 0.4, and set the maximum token length to a random value within the range of 600 to 1400 tokens.

Distinguishing features of our approach include the integration of "Character Role Play" and "Evolution of Thought" (EoT), the latter being akin to the AutoCoT methodology and introduced in Section \ref{sec:EoT}. "Character Role Play" broadens the exploration within the potent areas of the LLM's latent space, fostering innovative suggestions. EoT, on the other hand, introduces a feedback loop, refining the LLM's output over successive iterations for progressive improvement.

This methodology positions the GE framework not merely as a tool for random mutation but as a strategic guide for the LLM. It directs the exploration and enhancement of solutions in a more focused and effective manner, as detailed in our discussion of EoT in Section \ref{sec:EoT}.

\subsubsection{Character Role Play:}
\label{sec:character}

In the Guided Evolution framework, we approach the mutation process distinctively, diverging from conventional methods. Our initial experiments utilizing a LLM for code mutation identified a tendency towards producing standard solutions, exemplified by the frequent selection of a 0.2 dropout rate. This trend likely mirrors the commonality of such values in the model's training dataset, limiting the exploration of a more diverse solution space.

To address this, we implemented a method that incorporates some unconventional prompt templates, specifically designed to prompt the LLM away from generating these typical solutions. This method employs a range of randomly selected prompt templates, some of which guide the LLM to generate more atypical or novel code modifications. Descriptions of these prompt categories can be found in Table \ref{tab:prompt_cat}

\begin{table}
    \centering
    \begin{tabular}{p{2.0cm} p{5.7cm}}
        \hline
        \multicolumn{1}{c}{\textbf{Category}} & \multicolumn{1}{c}{\textbf{Summary}} \\
        \hline
        Hyperparam. & Modify existing parameter values to alter the model's behavior.
\\
        Hyperparam. Uncommon& Change parameter values to less conventional ones to explore diverse outcomes.
\\
        Complex& Introduce advanced functionality that could potentially enhance the model's accuracy.
\\
        Reduce Model Size& Streamline the code by reducing parameters with minimal impact on the model's performance.
\\
        Uncommon& Implement a distinctive or rarely used enhancement to the model.
\\
        Significant& Execute substantial modifications to the code, incorporating auxiliary functions for improved structure and capability.
\\
    \hline
    \end{tabular}
    \caption{Prompt Type Categories}
    \label{tab:prompt_cat}
\end{table}

This approach increased the diversity and scope of exploration in our preliminary tests, boosting the likelihood of developing solutions that surpass current state-of-the-art models. To further encourage diversity and enhance the quality of generated code, we introduced an element of “expert” roleplay in our prompts. This feature was intended to encourage the LLM to access its latent knowledge of “expert-level code” by casting it as a skilled AI researcher when queried. Namely, we incorporated a range of character personas for the LLM to adopt, each designed to induce different types and qualities of mutations. 

Specifically, we choose three unique character roles which act as compatible extensions to the previously mentioned 6 foundational prompts. These character types were chosen after manual analysis monitoring the quality of the produced models in response to both the standard prompts and the character-augmented prompts. Details regarding the three character roles, including their descriptions, are delineated in Table \ref{tab:characters}.

Our evolutionary algorithm utilizes six fundamental prompt templates, compatible with three distinct "expert" character roles. By combining these baseline prompts with the character role variations that modify the base template, we achieve a total of twenty-four diverse prompts. This variety in prompts fosters a broader and more enriched exploration of potential solutions by the LLM. 

\begin{table}
    \centering
    \begin{tabular}{p{1.5cm}p{6cm}}
        \hline
        \multicolumn{1}{c}{\textbf{Name}} & \multicolumn{1}{c}{\textbf{Summary Description}} \\
        \hline
        Expert& A top expert in machine learning with a deep understanding of advanced AI methods.
\\
        Dr. MaGoo& An AI innovator with a serendipitous approach, surprising peers with unorthodox model enhancements.
\\
        Innovative Scientist& A globally famous AI researcher known for creative and unconventional techniques.
\\
    \hline
    \end{tabular}
    \caption{Character Role Play}
    \label{tab:characters}
\end{table}

In future studies, prompt templates and character implementation are something that could be co-evolved as opposed to arbitrarily chosen.

\subsubsection{Evolution of Thought:}
\label{sec:EoT}

In 2022, Zhang et al \cite{zhang2022automatic}. pioneered the Auto Chain of Thought prompting (Auto-CoT), an innovative approach that automates the construction of demonstrations for LLMs. This method employs question clustering, where a dataset is divided into several clusters of similar questions. For each cluster, a representative question is selected, and its reasoning chain is generated using Zero-Shot-CoT with simple heuristics. Thus Auto-CoT addresses the limitations of manual and zero-shot CoT (Chain of Thought) prompting by leveraging the diversity of questions to mitigate the impact of errors in reasoning chains generated by LLMs. The key innovation lies in utilizing clustering algorithms to group similar questions, thus ensuring a variety of reasoning types and improving the robustness of the model's problem-solving capabilities. 

\begin{figure}
    \centering
    \includegraphics[width=1\linewidth]{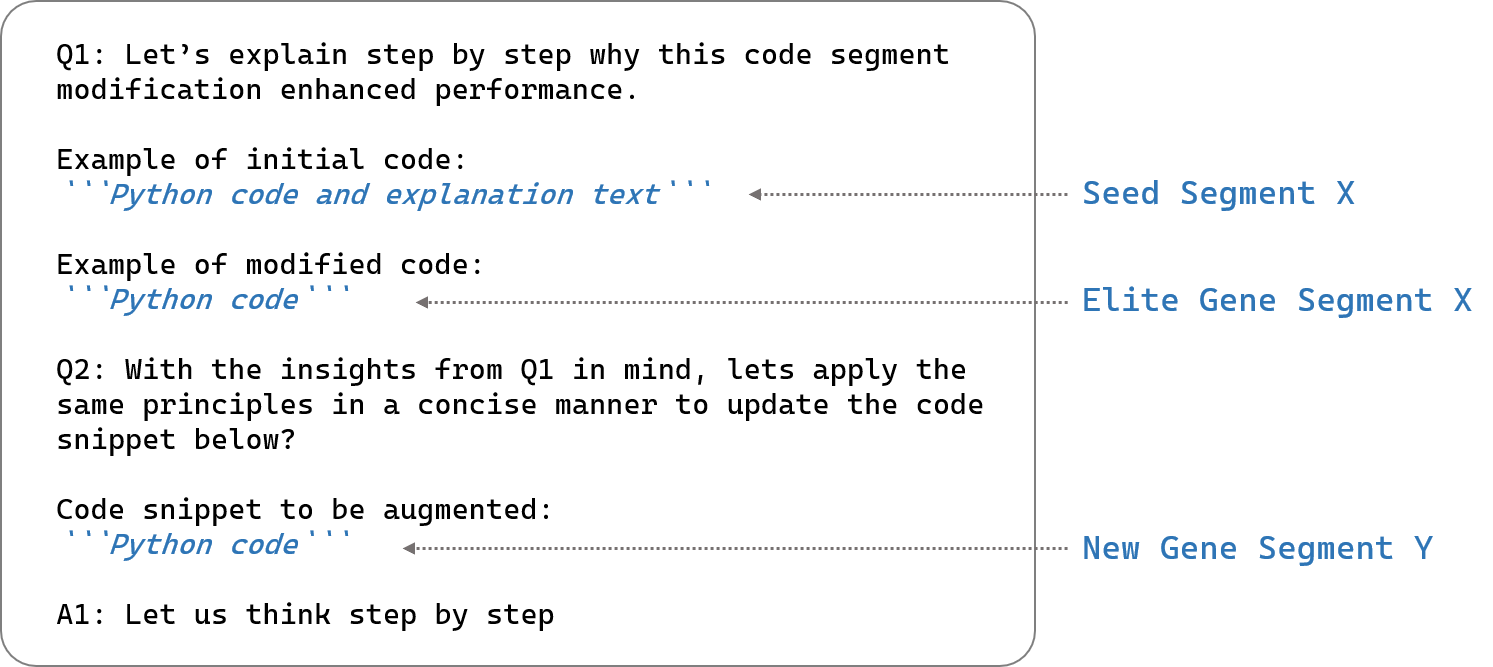}
    \caption{EoT prompt template}
    \label{fig:eot}
\end{figure}

Our study introduces the "Evolution of Thought" (EoT) methodology, which extends the reasoning principles of both Auto-CoT and Manual-CoT. What sets EoT apart is its incorporation of feedback from evolutionary processes within the Guided Evolution (GE) framework. In this framework, each mutation subsequent to the first generation can conduct a basic LLM mutation, “Character Role Play”, or the EoT mutation. EoT uniquely adopts the SPEA-2 elite selection algorithm to identify high-performing individuals from the previous generation. These individuals are then used as exemplary models for the LLM in subsequent mutations. This approach resonates with Zero-Shot-CoT through its “Let us think step by step” reasoning, and with Manual-CoT in its use of demonstrative examples. Leveraging the comparison of a selected elite block against its un-evolved seed serves as a thought guide for the LLM and encourages it to reflect about why mutations caused performance gains. The LLM is then asked to apply these derived insights to new code blocks, thus forming a performance enhancing feedback mechanism across generations. The EoT template is detailed in Figure \ref{fig:eot}.

Through analyzing empirical feedback on preceding concepts, EoT is designed to mimic and accelerate human like software development and invention.

\subsection{Evolutionary Loop}
This study introduces a novel approach in machine learning by integrating LLM adaptations into the three key components of evolutionary algorithms: genome creation, mutation, and mating. Detailed discussions on the adaptations of mating and mutation processes are presented in Section \ref{sec:mating} and \ref{sec:mutation}. The following section section provides an overview of the entire evolutionary process, which also covers how our method addresses gene creation.

In our evolutionary framework, genes are generated through LLM-driven mutations applied to copies of the state-of-the-art seed code. This gene creation method involves the seed code undergoing our unique LLM mutation procedure, intentionally excluding EoT mutations due to their reliance on previous change evaluations.

The evaluation process within this study is multi-objective, simultaneously considering accuracy on a holdout set and the model's parameter count. This dual-objective strategy enables the evolutionary algorithm to effectively balance accuracy and model size, optimizing both in tandem.

Elitism plays a crucial role in our approach due to the low likelihood of recovering or reconverging to a lost individual. We maintain elitism in the evolutionary loop through the Strength Pareto Evolutionary Algorithm 2 (SPEA-2) developed by Zitzler et al \cite{ZLTh_01_SPE}.  SPEA-2 was selected for its ability to preserve top-performing solutions while ensuring diversity. It achieves this balance via “fine-grained fitness assignment” and “k-th nearest neighbor density estimation” methods that enhance exploration and prevent premature convergence by evaluating solutions based on dominance relationships and density in the objective space.

In the context of selection for mating and mutation processes, the study leverages the NSGA-II algorithm, as delineated by Deb et al. \cite{996017}. This algorithm is characterized by its 'fast non-dominated sorting' technique coupled with 'crowding distance' assessments, pivotal for curating a diverse yet superior pool of solutions. The fast non-dominated sorting algorithm stratifies solutions by their dominance levels, while the crowding distance metric evaluates the population's spatial density, preferentially guiding the selection process towards sparser regions. Such a methodology ensures an optimal equilibrium between the caliber and the variety of solutions, a crucial factor for effective exploration within the multiobjective optimization domain. 

The integration of prompt variations significantly broadens the effective model search space, limited only by the LLM's vast output capabilities. This expansion greatly benefits the discovery of unique and effective solutions, but it also necessitates the retention of efficient individuals. To address this, before mutation and mating, both SPEA-2 and NSGA-II selections are conducted on populations of the same size. Post-augmentation and replacement through mating and mutation, the new individuals are concatenated with the SPEA-2 selected genes. Elite genes are only replaced if the new generation surpasses their performance. This strategy is employed to mitigate issues arising from the considerably expansive search space , ensuring the capture and retention of optimal solutions.

\begin{algorithm}
\DontPrintSemicolon
\caption{LLM-Guided Evolution Algorithm}

\SetKwFunction{FMain}{LLMGuidedEvolution}
\SetKwProg{Fn}{Function}{:}{}
\Fn{\FMain{$SeedCode$}}{
    $Population \gets \text{InitializePopulation}(SeedCode)$\;
    $ElitePopulation \gets \emptyset$\;
    $HallOfFame \gets \emptyset$\;
    \While{not \text{ConvergenceCondition}()}{
        $EvaluatedPopulation \gets \text{Evaluate}(Population)$\;
        $EliteCandidates \gets \text{SPEA2}(EvaluatedPopulation)$\;
        $SelectionForMating \gets \text{NSGA2}(EvaluatedPopulation)$\;
        $MatedPopulation \gets \text{LLMMate}(SelectionForMating)$\;
        $MutatedPopulation \gets \text{LLMMutate}(MatedPopulation)$\;
        $HallOfFame \gets \text{UpateHoF}(EvaluatedOffspring)$\;
        
        $CombinedPopulation \gets \text{Concatenate}(EliteCandidates, EvaluatedOffspring)$\;
        $Population \gets CombinedPopulation$\;
    }
    \KwRet $HallOfFame$\;
}
\end{algorithm}

In our mutation process, a probabilistic method, directed by "prob\_eot" is used to alternate randomly between fixed prompt mutation and "Evolution of Thought" (EoT). When fixed prompt mutation is selected, the system chooses a single prompt template from either traditional direction prompts or Character Role Play prompts, as outlined in Section \ref{sec:character}. 

\begin{algorithm}
\DontPrintSemicolon
\caption{Detailed LLM Mutation Process}

\SetKwFunction{FMutation}{LLMMutate}
\SetKwProg{Fn}{Function}{:}{}
\Fn{\FMutation{$Gene$}}{
    \tcp{Select a random code block Y from the gene's alternative blocks}
    $BlockY \gets \text{SelectRandomBlock}(Gene)$
    
    \tcp{Probabilistic method to choose mutation type}
    \eIf{\text{Random}() < prob\_eot}{
        \tcp{Perform Fixed Prompt Mutation}
        $Prompt \gets \text{SelectRandomPrompt}(\text{FixedPrompts} \cup \text{RolePlayPrompts})$
        
        $MutatedBlockY \gets \text{FixedPromptMutation}(BlockY, Prompt)$
    }{
        \tcp{Perform Evolution of Thought (EoT)}
        $Elite \gets \text{SPEA2Selection}(LastGeneration, k)$
        $EliteIndividual \gets \text{SelectRandomIndividual}(Elite)$
        $(BlockX_{Elite}, BlockX_{Seed}) \gets \text{SelectRandomBlock}(EliteIndividual \cup SeedCode)$
        $MutatedBlockY \gets \text{EoTMutation}(BlockY, BlockX_{Elite}, BlockX_{Seed})$
    }
    
    \tcp{Replace the original block Y with the mutated block Y}
    $Gene \gets \text{ReplaceBlock}(Gene, BlockY, MutatedBlockY)$
    
    \KwRet $Gene$
}
\end{algorithm}

The developed evolutionary loop aims to optimally utilize the explorative capabilities of LLMs while ensuring to exploit effective adaptations. 

\subsection{Example Problem}
In the pursuit of advancing neural architecture search methodologies, our investigation employed the CIFAR10 dataset, a seminal dataset developed by the Canadian Institute for Advanced Research, as a primary benchmark \cite{krizhevsky2009learning}. The dataset encompasses 60,000 32x32 pixel color images across 10 classes, with a standard division of 50,000 images for training and 10,000 for testing.

Our rationale for selecting CIFAR10 was twofold: firstly, the challenge presented by its low-resolution images, which necessitates sophisticated object recognition algorithms to address limited detail availability; and secondly, its extensive usage in the machine learning domain, which has resulted in a proliferation of highly effective SOTA models to benchmark our auto-enhancing GE against.

CIFAR10's balance of complexity and computational feasibility makes it an ideal candidate for evaluating new neural architectures. It demands intricate learning algorithms for accurate classification while remaining manageable in terms of data handling and computational requirements, a crucial consideration for developing from scratch a novel NAS methodology such as Guided Evolution.

Conclusively, the CIFAR10 dataset's historical significance in machine learning, combined with its unique challenges and practical usability, positions it as an optimal choice for our research. It not only serves as a fundamental benchmark but also as a catalyst for designing a new and effective NAS paradigm in a demanding and well-established domain.

\subsection{State of the Art}
Our study aimed to assess our framework's capacity to outperform human-engineered designs in well-researched domains. We used a state-of-the-art CIFAR-10 classifier as our seed model for autonomous evolution, chosen for its complex block and layer diversity, presenting a challenging environment for evolution. 

A pivotal aspect of our methodology was the deliberate choice of models with fewer parameters. This decision was twofold: firstly, models with a lower parameter count emphasize the critical role of architectural ingenuity. Secondly, such models are more conducive to exhaustive experimentation within limited time frames, ensuring thorough evaluation.

We opted for "ExquisiteNetV2", a model innovatively crafted by Zhou and Su in 2022  \cite{zhou2022novel}, as our foundational model. ExquisiteNetV2 is not only compact, ranking ninth in terms of size on the CIFAR-10 benchmark, but its architecture also comprises a variety of effective blocks. This aligns seamlessly with our goal of advancing sophisticated architectures through our evolutionary framework. 

ExquisiteNetV2 is characterized by several innovative components. It incorporates Extreme-Value-Expansion (EVE) Blocks, which concatenate min and max pooling layers to rapidly discard unimportant features while retaining critical ones. The Feature-Concentrator Block integrates a 4x4 depthwise convolution with the EVE block to focus on preserving raw image features. In an effort to reduce parameter count, SE-Layer-Normalization (SE-LN) Blocks replace fully-connected layers with Layer Normalization in a Squeeze Excitation blocks to further reduce parameter count.  The DFSEBV2 Block, a cornerstone of the ExquisiteNetV2 architecture, integrates pointwise convolutions, batch normalization, and depthwise convolutions for efficient feature processing. It uses SiLU and Hardswish activation functions for advanced pattern learning, and includes either an SE-LN or a standard SE block for improved channel-wise feature recalibration. This block's design is further enhanced by residual connections, ensuring optimal learning efficiency and computational economy. These elements make ExquisiteNetV2 a fitting candidate to demonstrate the capabilities of our guided evolutionary framework.

\section{Results}

In this investigation, our primary objective was to leverage LLM Guided Evolution for the generation of ExquisiteNetV2 variants, with the dual goals of maintaining high test accuracy while constraining model size. Furthermore, this research endeavored to evaluate the contributions of two novel methodologies: Evolution of Thought (EoT) and Character Role Play (CRP), within the context of this evolutionary framework.

The ensuing sections will delve into the specifics of our findings. Section \ref{sec:evo_seeds} will detail the performance metrics of the most effective models derived from this study. Subsequent to this, Section \ref{sec:ablation} will present an ablation study that explores the individual and combined influences of the EoT and CRP methodologies. This latter section aims to offer observations regarding their respective impact on the GE framework.

\subsubsection{Evolved Seeds:}
\label{sec:evo_seeds}

The experimental results from our Guided Evolution approach are promising. The autonomous framework has successfully evolved ExquisiteNetV2 variants, surpassing the original model's benchmarks of 92.52\% accuracy and a parameter count of 0.518230M.

\begin{figure}
    \centering
    \includegraphics[width=1\linewidth]{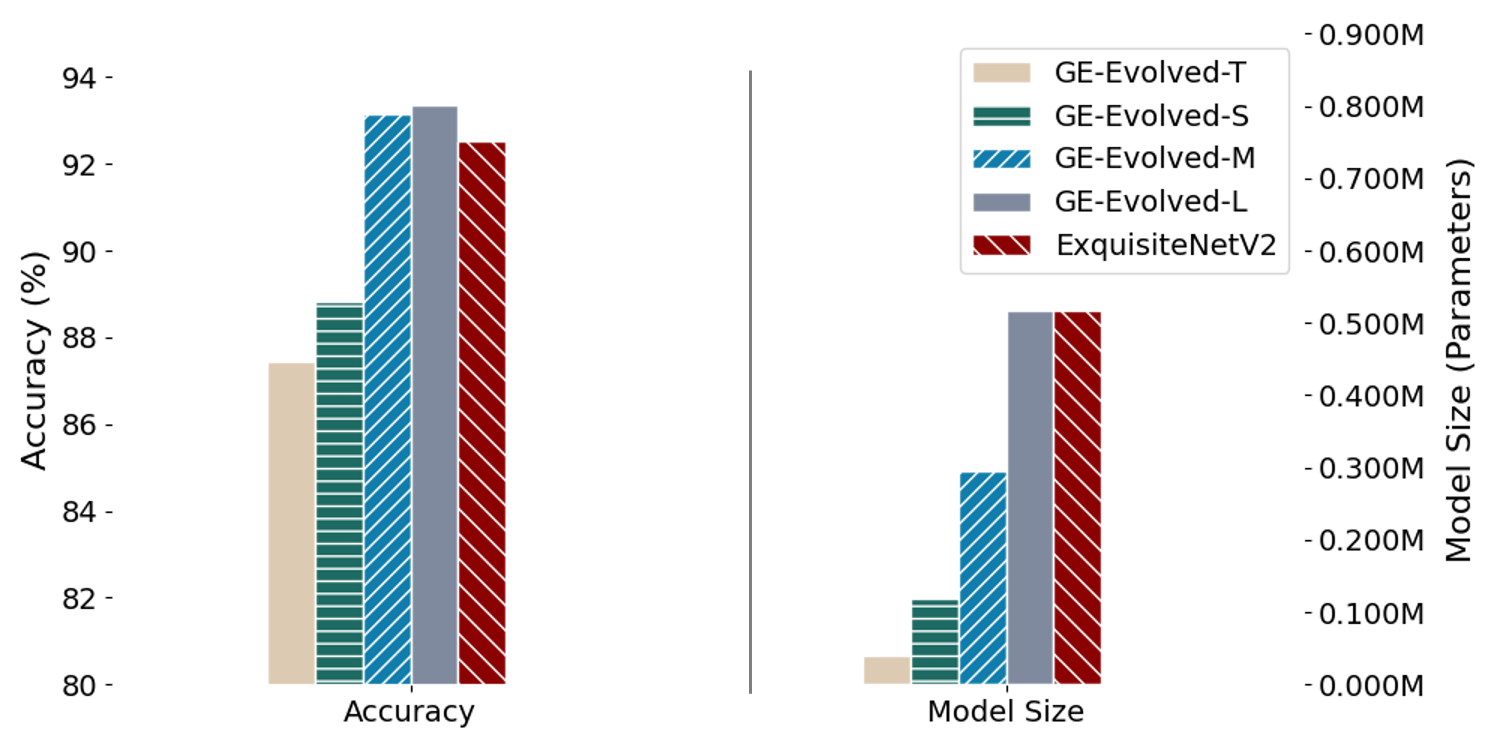}
    \caption{Metrics of evolved ExquisiteNetV2 models.}
    \label{fig:best_models}
\end{figure}

A notable example is the "GE-Evolved-L" variant, achieving an accuracy of 93.34\% with a parameter count of 0.518230M, aligning exactly with the original ExquisiteNetV2's size. Figure \ref{fig:best_models} provides a comparative illustration of this variant against the seed model, illustrating the capacity of our GE framework to autonomously enhance neural architectures with human-like expertise.

Moreover, significant progress was made in enhancing parameter efficiency. For instance, the "GE-Evolved-M" variant achieved a remarkable 43.1\% reduction in parameter count compared to ExquisiteNetV2, alongside an improved test accuracy of 93.16\%. This efficiency and performance are visually represented in Figure \ref{fig:best_models}. Additionally, the "GE-Evolved-S" and "GE-Evolved-T" variants, while slightly less accurate than the seed model at 88.83\% and 87.45\% respectively, showcased a dramatic improvement in model size to 0.119254M and 0.039190M, illustrating a substantial advancement in model compactness without significant accuracy trade-offs.
 
\begin{figure}
    \centering
    \includegraphics[width=1\linewidth]{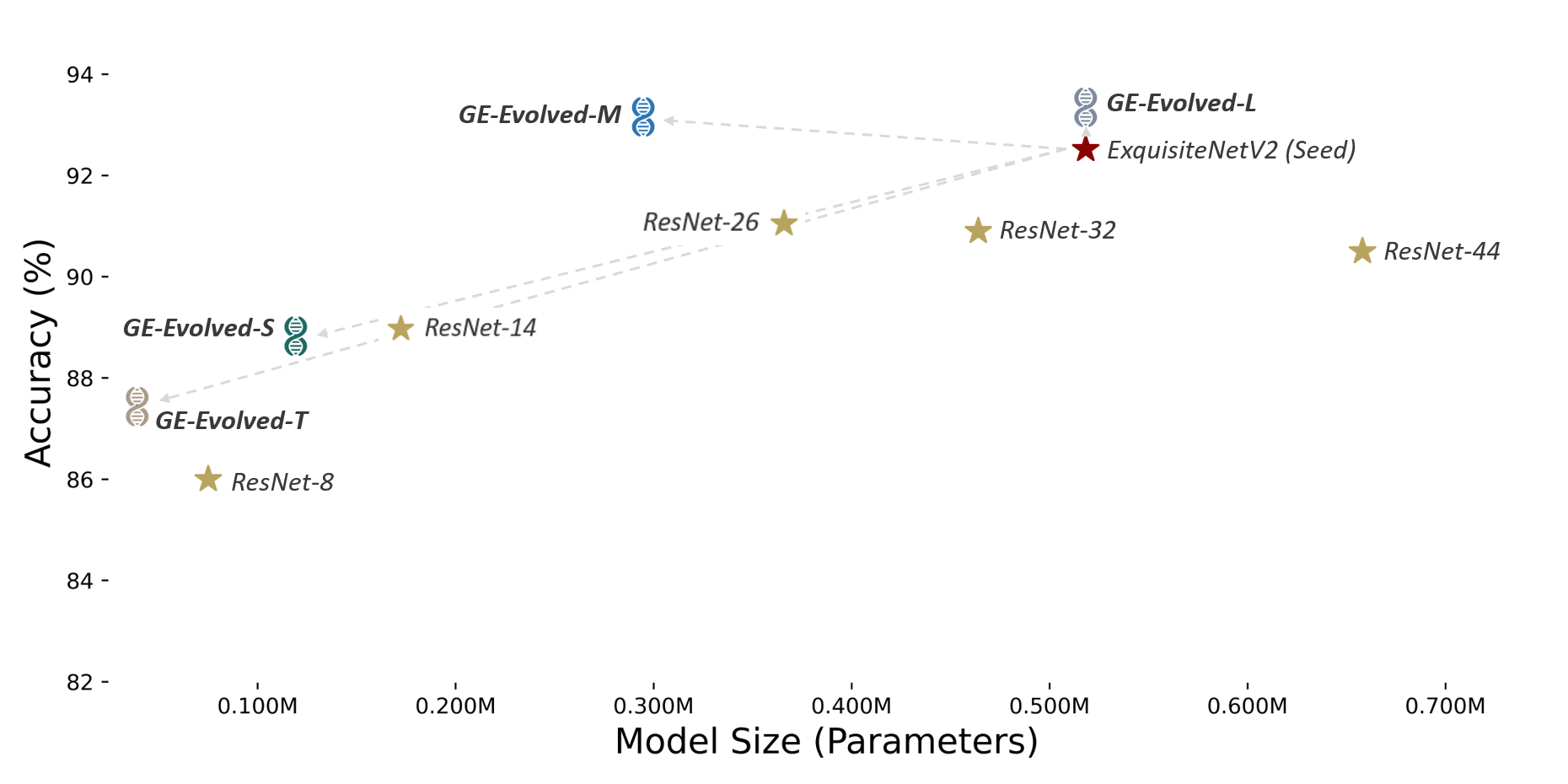}
    \caption{Evolved ExquisiteNetV2 models compared to State of the Art.}
    \label{fig:sota_plot}
\end{figure}

To illustrate the performance improvements achieved through Guided Evolution, we compare ExquisiteNetV2's enhanced performance with other leading low-parameter CIFAR-10 classifiers in Figure \ref{fig:sota_plot}. This comparison provides a contextual understanding of the advancements made by our autonomous framework. Looking ahead, we plan to expand the application of the GE methodology to a broader range of seed models in future research endeavors.

\begin{figure}
    \centering
    \includegraphics[width=1\linewidth]{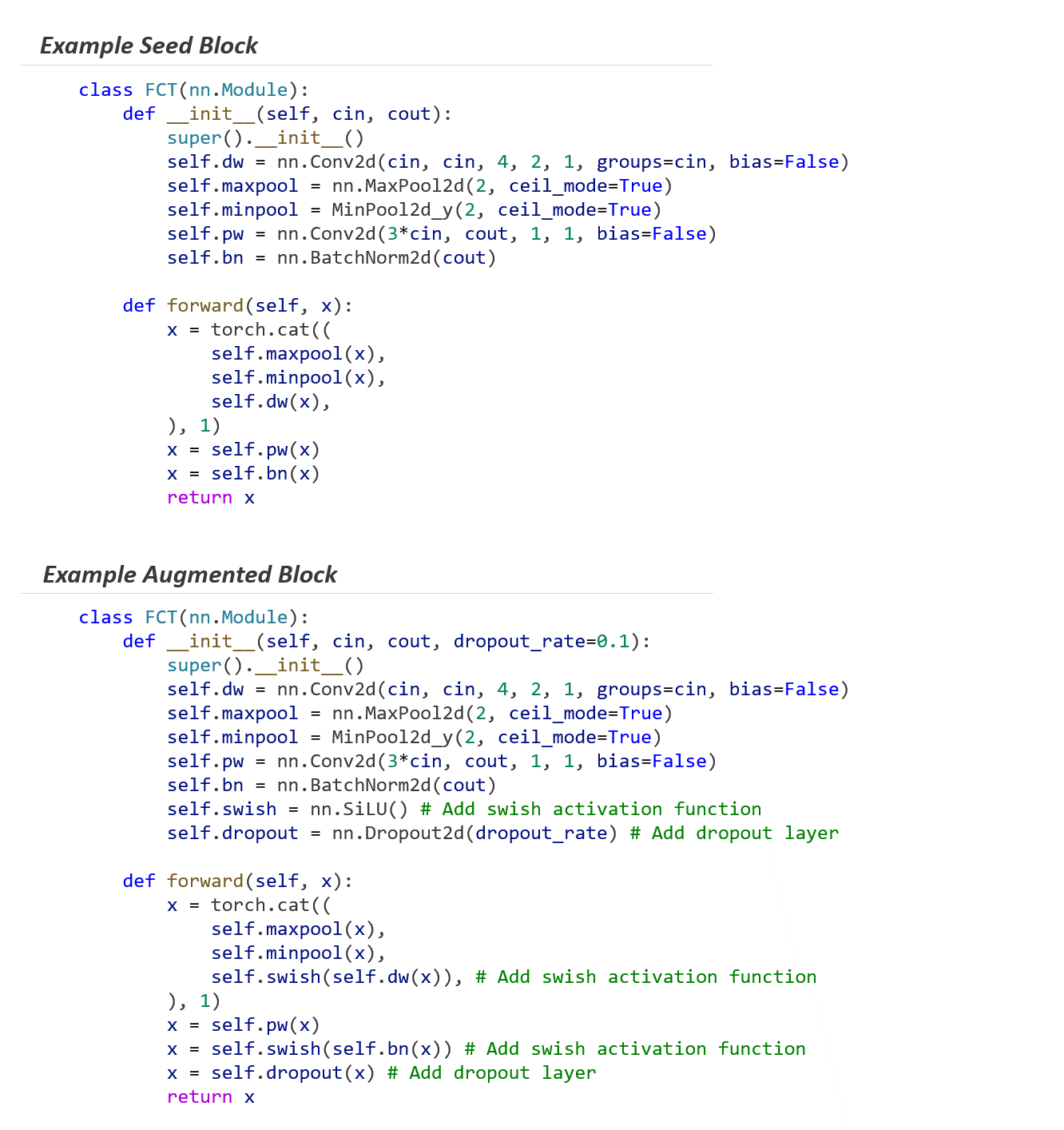}
    \caption{Example of an augmented code block.}
    \label{fig:code_change}
\end{figure}

To illustrate the types of augmentations produced by the Genetic Engineering (GE) process, Figure \ref{fig:code_change} showcases an example of an augmented code block from an effective individual (GE-Evolved-M). 

In addition to the notable performance improvements observed in this study, the GE framework also demonstrated potential for further advancement over longer evolutionary periods. This extended potential can likely be attributed to the vast array of design pathways that the LLM-guided framework is capable of exploring.

It is important to highlight that the cutoff point of the evolutionary process in this study was due to time constraints and not because a significant convergence point was reached.

\subsubsection{Ablation Study:}
\label{sec:ablation}

\begin{figure}
    \centering
    \includegraphics[width=1\linewidth]{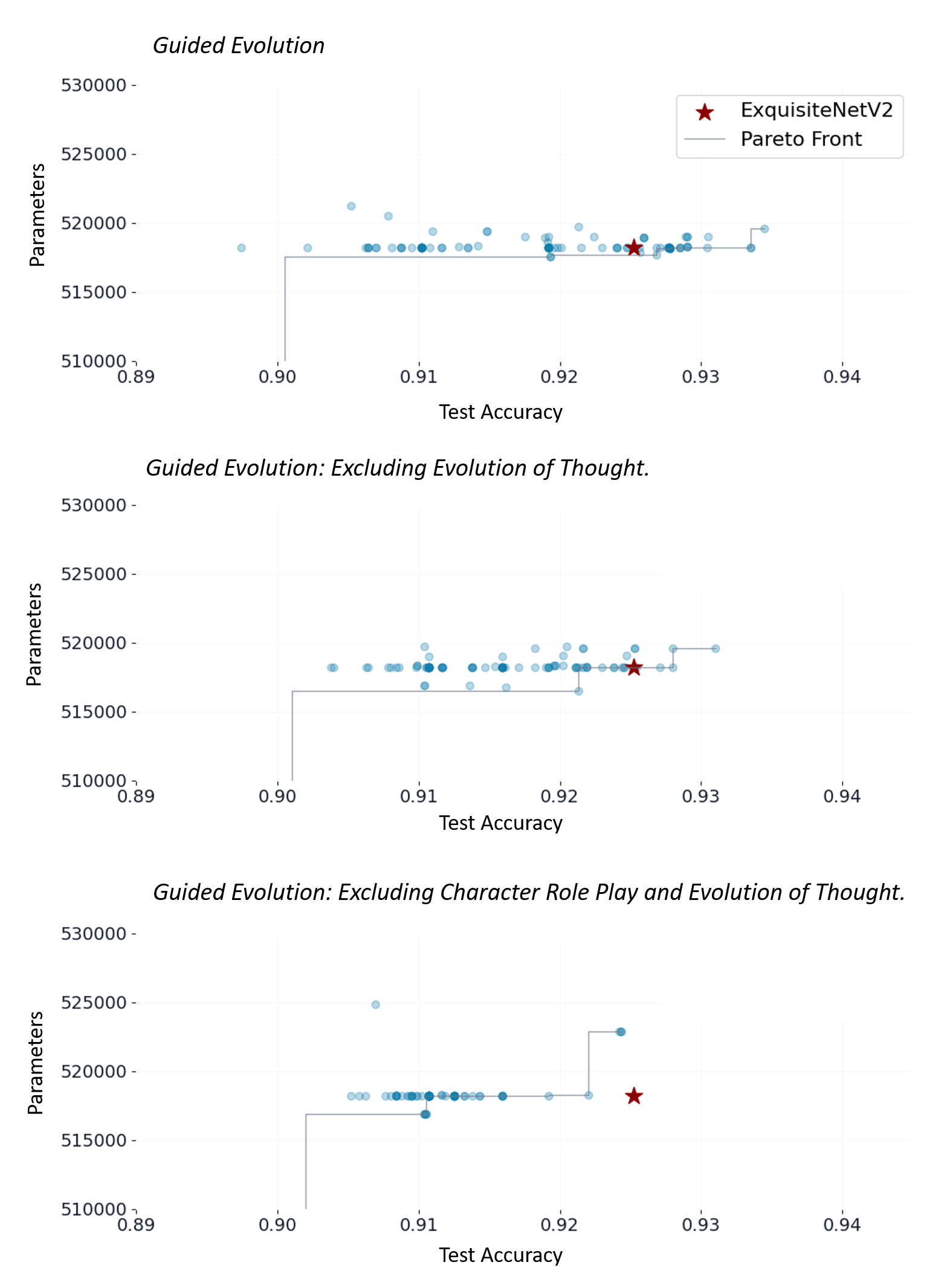}
    \caption{Pareto Frontier of Guided Evolution variants: Generation 9}
    \label{fig:generation}
\end{figure}

A pivotal aspect of our study was assessing the impact of Evolution of Thought (EoT) and Character Role Play on the Guided Evolution process. To this end, we conducted variant experiments: one excluding EoT and another excluding both EoT and Character Role Play. The results demonstrated that the incorporation of EoT and Character Role Play substantially influenced the evolution trajectory. Notably, the inclusion of EoT accelerated the convergence towards optimal solutions. This enhancement is graphically represented through the comparison of Pareto frontiers at generation 9, as illustrated in Figure \ref{fig:generation}.

\begin{figure}
    \centering
    \includegraphics[width=1\linewidth]{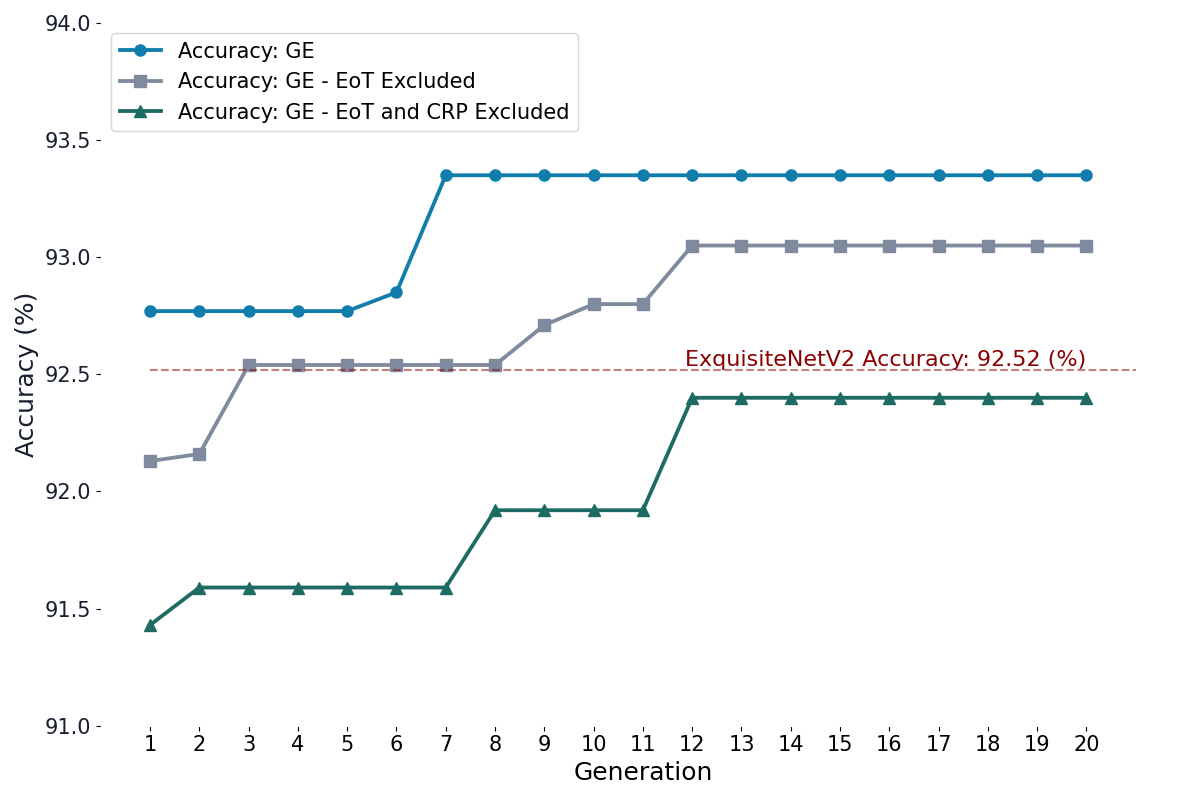}
    \caption{Guided Evolution exclusionary study.}
    \label{fig:abl}
\end{figure}

To further elucidate the development trajectory of individuals that conformed to our primary objective—enhancing accuracy without increasing the parameter count beyond that of ExquisiteNetV2—we tracked the progression of the most accurate model at each generation within the three distinct Guided Evolution runs that maintained the same or fewer parameters than ExquisiteNetV2. The models that ultimately achieved the highest accuracy, while adhering to the parameter constraints of ExquisiteNetV2, are highlighted in Figure \ref{fig:abl}. This figure depicts the influence of the EoT and Character Role Play methodologies on the trajectory and efficiency of the Guided Evolution process.

Our findings show that the GE framework quickly identified model overfitting. Implementing various regularization methods early on led to significant improvements. While crossover facilitated horizontal transfer of updates across gene blocks, the EoT feedback mechanism enabled vertical propagation of these changes. Thus, if adding regularization to an elite example's code block X improved performance, EoT could extend these updates to another genome's block Y. Consequently, EoT ensures intrinsic feedback and synchronicity across a genome's disparate code blocks.

\section{CONCLUSION}

In this study, we introduced Guided Evolution (GE), a novel methodology for harnessing the power of Large Language Models (LLMs) in advancing neural architectures. Central to GE are two innovative concepts: Evolution of Thought (EoT) and Character Role Play (CRP), which significantly contribute to the intelligent evolution of model architectures and infuse diversity and creativity into the evolutionary process, respectively.

Our findings, especially with models such as "GE-Evolved-L", "GE-Evolved-M", "GE-Evolved-S" and "GE-Evolved-T" demonstrate GE's potential in autonomously enhancing state-of-the-art neural architectures. These outcomes underline the efficacy of EoT and CRP within the GE framework.

It's important to note that our ablation studies, conducted with a single trial for each variant due to time constraints, serve as an initial exploration. Future work is needed to fully assess the impact of EoT and CRP within this algorithm framework.

Future research will aim to expand GE across various datasets and domains, fully harnessing its capabilities in machine learning. We will explore the impact of GE on the efficiency of Neural Architecture Search (NAS), the role of instructive commenting in code gene sequences, and the effects of different LLM models on GE's parameter space exploration.

In conclusion, our study lays a foundational framework for future exploration in GE, with the anticipation that advancements in NLP and LLM development will significantly enhance Guided Evolution's performance. This work opens new horizons for the exploitation of LLMs in genetic algorithms and multi-objective optimization, promising an exciting future for models that advance models. For access to our code, datasets, and supplementary materials, visit our GitHub repository: \url{https://github.com/clint-kristopher-morris/llm-guided-evolution}. We invite the community to explore, replicate, and contribute to GE's development.

\section{ACKNOWLEDGEMENTS}

We thank Aaron McDaniel for his expert guidance on genetic algorithms, which was pivotal to our research's success. His insights significantly enriched our approach, demonstrating both depth and practical applicability. 


\bibliographystyle{ACM-Reference-Format}
\bibliography{sample-authordraft}


\end{document}